\documentclass[11pt,a4paper]{article}
\usepackage[hyperref]{eacl2021}
\usepackage{times}
\usepackage{latexsym}
\usepackage{times}
\usepackage{latexsym}

\usepackage{microtype}

\usepackage{hyperref}
\usepackage{url}
\usepackage{graphicx}
\usepackage{enumitem}
\usepackage{multirow}
\usepackage{booktabs}
\usepackage{commath}
\usepackage{bbm}
\usepackage{amsmath}
\usepackage{amssymb}
\usepackage{pifont}% http://ctan.org/pkg/pifont
\usepackage{xcolor}
\usepackage{makecell, cellspace, caption}
\usepackage[rightcaption]{sidecap}

\aclfinalcopy % Uncomment this line for the final submission
\def\aclpaperid{1059} %  Enter the acl Paper ID here

\title{{\color{black}Extremely Small BERT Models from Mixed-Vocabulary Training}}

\author{Sanqiang Zhao*$^1$,$\ $ Raghav Gupta*$^2$,$\ $ Yang Song$^3$, \and Denny Zhou$^4$\\[2ex]
$^1$University of Pittsburgh, Pittsburgh, PA$\quad$ \texttt{sanqiang.zhao@pitt.edu}\\%
    $^2$Google Research, Mountain View, CA$\quad$ \texttt{raghavgupta@google.com}\\
    $^3$Kuaishou Technology, Beijing, China$\quad$ \texttt{yangsong@kuaishou.com}\\
    $^4$Google Brain, Mountain View, CA$\quad$ \texttt{dennyzhou@google.com}
    }

\newcommand{\RNum}[1]{\uppercase\expandafter{\romannumeral #1\relax}}

\begin{document}

\maketitle

\begin{abstract}

Pretrained language models like BERT have achieved good results on NLP tasks, but are impractical on resource-limited devices due to memory footprint. A large fraction of this footprint comes from the input embeddings with large input vocabulary and embedding dimensions. Existing knowledge distillation methods used for model compression cannot be directly applied to train student models with reduced vocabulary sizes. To this end, we propose a distillation method to align the teacher and student embeddings via mixed-vocabulary training. Our method compresses \mbox{\color{black}BERT\textsubscript{LARGE}} to a task-agnostic model with smaller vocabulary and hidden dimensions, which is an order of magnitude smaller than other distilled BERT models and offers a better size-accuracy trade-off on language understanding benchmarks as well as a practical dialogue task.

\end{abstract}

\section{Introduction}

Recently, pre-trained context-aware language models like ELMo \citep{peters2018deep}, GPT \citep{radford2019language}, BERT \citep{devlin2018bert} and XLNet \citep{yang2019xlnet} have outperformed traditional word embedding models like Word2Vec \citep{mikolov2013distributed} and GloVe \citep{pennington2014glove}, and achieved strong results on a number of language understanding tasks. However, these models are typically too huge to host on mobile/edge devices, especially for real-time inference. Recent work has explored, inter alia, knowledge distillation \citep{ba2014deep, hinton2015distilling} to train small-footprint student models by implicit transfer of knowledge from a teacher model.

\let\svthefootnote\thefootnote
\let\thefootnote\relax\footnote{\noindent Asterisk (\textbf{*}) denotes equal contribution. Research conducted when all authors were at Google.}
\addtocounter{footnote}{-1}
\let\thefootnote\svthefootnote

% incl. natural language inference \citep{williams2018broad} and question answering \citep{rajpurkar2016squad,lai2017race}

Most distillation methods, however, need the student and teacher output spaces to be aligned. This complicates task-agnostic distillation of BERT to smaller-vocabulary student BERT models since the input vocabulary is also the output space for the masked language modeling (MLM) task used in BERT. This in turn limits these distillation methods' ability to compress the input embedding matrix, that makes up a major proportion of model parameters e.g. the $\sim$30K input WordPiece embeddings of the BERT\textsubscript{BASE} model make up over $21\%$ of the model size. This proportion is even higher for most distilled BERT models, owing to these distilled models typically having fewer layers than their teacher BERT counterparts.

We present a task and model-agnostic distillation approach for training small, reduced-vocabulary BERT models running into a few megabytes. In our setup, the teacher and student models have incompatible vocabularies and tokenizations for the same sequence. We therefore align the student and teacher WordPiece embeddings by training the teacher on the MLM task with a mix of teacher-tokenized and student-tokenized words in a sequence, and then using these student embeddings to train smaller student models. Using our method, we train compact 6 and 12-layer reduced-vocabulary student models which achieve competitive performance in addition to high compression for benchmark datasets as well as a real-world application in language understanding for dialogue.

\section{Related Work}

\begin{figure*}[t!]
\begin{center}
\includegraphics[width=1.00\linewidth]{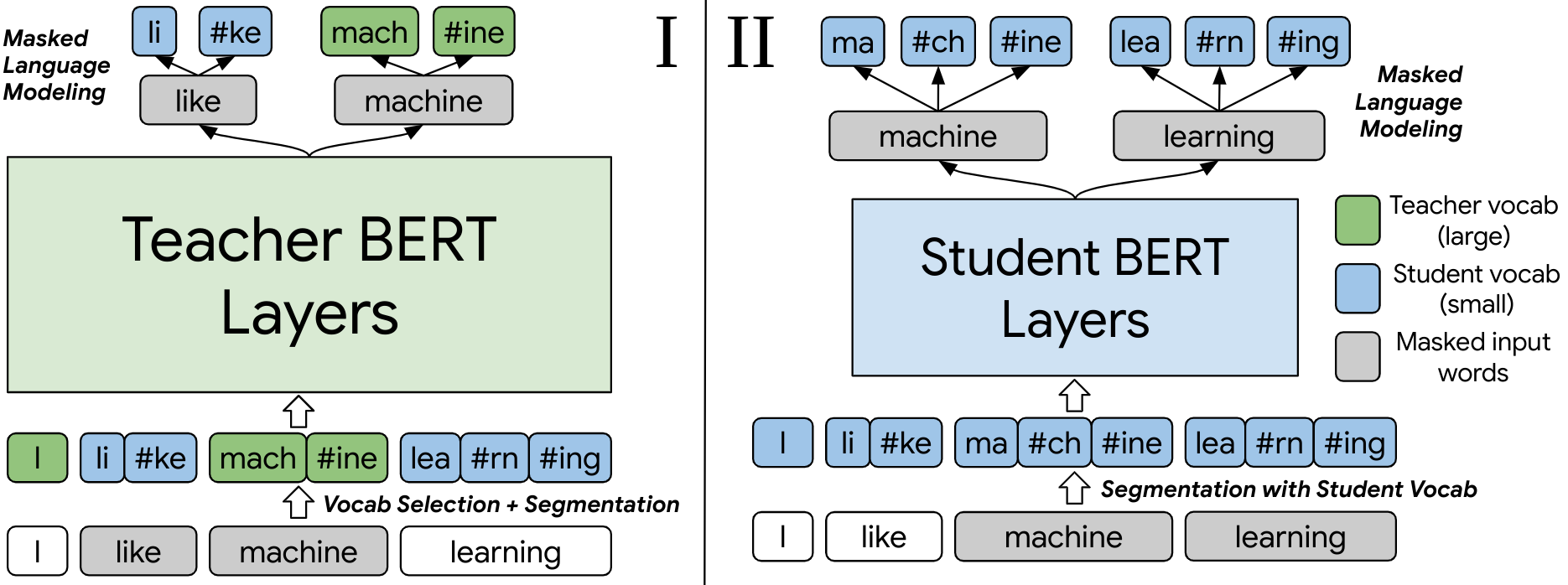}
% The figure lives at https://docs.google.com/drawings/d/1MrrFDf11zRwQD8N6scxcQddRsij5KCSkxeBNjPbq7Cg/edit?usp=sharing
\end{center}
\caption{Depiction of our {\color{black}mixed-vocabulary training}  approach. (Left) Stage \RNum{1} involving retrained teacher BERT with default config (e.g., 30K vocabulary, 768 hidden dim) and mixed-vocabulary input. (Right) Stage \RNum{2} involving student model with smaller vocabulary (5K) and hidden dims (e.g., {\color{black}256}) and embeddings initialized from stage \RNum{1}.}
\label{fig:model}
\end{figure*}

Work in NLP model compression falls broadly into four classes: matrix approximation, weight quantization, pruning/sharing, and knowledge distillation.

The former two seek to map model parameters to low-rank approximations \citep{tulloch2017high} and lower-precision integers/floats \citep{chen2015compressing, zhou2018adaptive, shen2019q} respectively. In contrast, pruning aims to remove/share redundant model weights \citep{li2016pruning, lan2019albert}. More recently, dropout \cite{srivastava2014dropout} has been used to cut inference latency by early exit \cite{fan2019reducing,xin2020deebert}.

% Lowrank sindhwani2015structured, quant lin2017towards, pruning anwar2017structured

% \begin{table}[h]
% \small
% \centering
% \setlength\tabcolsep{1.0pt}
% \def\arraystretch{1.5}
% \resizebox{\columnwidth}{!}{
% \begin{tabular}{lccc}
% \toprule
% \textbf{Model} & Min Params   & Task-agnostic   & Model-agnostic\\ 
% \hline
% PKD \cite{sun2019patient} & 45.7M & $\checkmark$ & $\times$   \\
% TinyBERT \cite{jiao2019tinybert} & 14.5M & $\times$  & $\times$   \\
% DistilBERT \cite{sanh2019distilbert} & 52.2M & $\times$ & $\times$    \\
% MobileBERT \cite{mobilebert} & 25.3M & $\checkmark$ & $\times$    \\
% BERT-of-Theseus \citep{xu2020bert} & 66.1M & $\checkmark$ & $\checkmark$ \\ 
% DualTrain (Ours) & {\color{black}\textbf{62x}} & $\checkmark$ & $\checkmark$\\
% \toprule
% \end{tabular}
% }
% \caption{Comparison of BERT distillation methods.}
% \label{tab:compare_bert}
% \end{table}

Knowledge distillation focuses on implicit transfer of knowledge as soft teacher predictions \citep{tang2019distilling}, attention distributions \citep{zagoruyko2016paying} and intermediate outputs \citep{romero2014fitnets}. Approaches close to our work rely on similar methods \citep{sanh2019distilbert, sun2019patient}, while others involve combinations of layer-wise transfer \cite{mobilebert}, task-specific distillation  \cite{jiao2019tinybert}, architecture search \citep{chen2020adabert} and layer dropout \cite{xu2020bert}; many of these are specific to the transformer layer \cite{vaswani2017attention}.

Another highly relevant line of work focuses on reducing the size of the embedding matrix, either via factorization \citep{shu2018compressing, lan2019albert} or vocabulary selection/pruning \citep{provilkov2019bpe, chen2019large}.
% yim2017gift

% We compare several such approaches to BERT in Table  \ref{tab:compare_bert}. However, these do not apply to our setup due to unaligned teacher and student vocabularies. % , as are vocab selection methods \citep{chen2019large} since BERT models are costly to train repeatedly.

% teaching a student model to match soft output label distributions from a larger model alongside the ground-truth distribution and works well for many tasks incl. machine translation \citep{kim2016sequence} and language modeling \citep{yu2018device}.

% Not limited to the teacher model outputs, distillation may be performed via attention transfer \citep{zagoruyko2016paying}, or feature maps and intermediate model outputs \citep{romero2014fitnets, yim2017gift}. 

% \citep{joulin2016fasttext}

\section{Proposed Approach}
Here, we discuss our rationale behind reducing the student vocabulary size and its challenges, followed by our {\color{black}\textit{mixed-vocabulary}} distillation approach.

\subsection{Student Vocabulary}

WordPiece (WP) tokens \citep{wu2016google} are subword units obtained by applying greedy segmentation to a training corpus. Given such a corpus and a number of desired tokens $D$, a WordPiece vocabulary is generated by selecting $D$ subword tokens such that the resulting corpus is minimal in the number of WordPiece when segmented according to the chosen WordPiece model. The greedy algorithm for this optimization problem is described in more detail in \citet{sennrich2016neural}. Most published BERT models use a vocabulary of 30522 WordPieces, obtained by running the above algorithm on the Wikipedia and BooksCorpus \citep{zhu2015aligning} corpora with a desired vocabulary size $D$ of 30000. 

For our student model, we chose a target vocabulary size $D$ of 5000 WordPiece tokens. Using the same WordPiece vocabulary generation algorithm and corpus as above, we obtain a 4928-WordPiece vocabulary for the student model. This student vocabulary includes all ASCII characters as separate tokens, ensuring no out-of-vocabulary words upon tokenization with this vocabulary. Additionally, the 30K teacher BERT vocabulary includes $93.9\%$ of the WP tokens in this 5K student vocabulary but does not subsume it. We explore other strategies to obtain a small student vocabulary in Section \ref{section:discussion}.

For task-agnostic student models, we reuse BERT's masked language modeling (MLM) task: words in context are randomly masked and predicted given the context via softmax over the model's WP vocabulary. Thus, the output spaces for our teacher (30K) and student (5K) models are unaligned. This, coupled with both vocabularies tokenizing the same words differently, means existing distillation methods do not apply to our setting.

\subsection{Mixed-vocabulary training}
\label{sec:dual_training}
We propose a two-stage approach for implicit transfer of knowledge to the student via the student embeddings, as described below.\\[7pt]
\noindent \textbf{Stage \RNum{1} (Student Embedding Initialization):} We first train the student embeddings with the teacher model initialized from BERT\textsubscript{LARGE}. For a given input sequence, we mix the vocabularies by randomly selecting (with probability $p_{SV}$, a hyperparameter) words from the sequence to segment using the student vocabulary, with the other words segmented using the teacher vocabulary. As in Figure \ref{fig:model} on the left, for input  [`\textit{I}', `\textit{like}', `\textit{machine}', `\textit{learning}'], the words   \textit{`like'} and \textit{`learning'} are segmented using the student vocabulary (in blue), with the others using the teacher vocabulary (in green). Similar to \citet{lample2019cross}, this step seeks to align the student and teacher embeddings for the same tokens: the model learns to predict student tokens using context which is segmented using the teacher vocabulary, and vice versa. 

Note that since the student embeddings are set to a lower dimension than the teacher embeddings, as they are meant to be used in the smaller student model, we project the student embeddings up to the teacher embedding dimension using a trainable affine layer before these are input to the teacher BERT. We choose to keep the two embedding matrices separate despite the high token overlap: this is partly to keep our approach robust to lower vocabulary overlap settings, and partly due to empirical considerations described in Section \ref{section:discussion}.

Let $\theta_s/eb_s$ and $\theta_t/eb_t$ denote the transformer layer and embedding weights for the student and teacher models respectively. The loss defined in Equation \ref{eq:stage1_loss} is the MLM cross entropy summed over masked positions $M_t$ in the teacher input. $y_i$ and $c_i$ denote the predicted and true tokens at position $i$ respectively and can belong to either vocabulary. $v_i {\in} {\{s{,}t\}}$ denotes the vocabulary used to segment this token. Separate softmax layers $P_{v_i}$ are used for token prediction, one for each vocabulary, depending on the segmenting vocabulary $v_i$ for token $i$. All teacher parameters ($\theta_t$, $eb_t$) and student embeddings ($eb_s$) are updated in this step.
\begin{equation}
\resizebox{.87\hsize}{!}{$L_{s_1} = {-}\sum_{i\in M_t} ({\log} P_{v_i}(y_i{=}c_i|\theta_{t}, eb_s, eb_t))$}
\label{eq:stage1_loss}
\end{equation}
\noindent \textbf{Stage \RNum{2} (Student Model Layers):}
{\color{black}With student embeddings initialized in stage \RNum{1}, we now train the student model normally i.e., using only the student vocabulary and discarding the teacher model}. Equation \ref{eq:final_loss} shows the student MLM loss where $M_s$ is the set of positions masked in the student input. All student model parameters ($\theta_s$, $eb_s$) are updated.
\begin{equation}
\resizebox{.85\hsize}{!}{$L_{s_2} = {-}\sum_{i\in M_s} {\log} P_{s}(y_i{=}c_i|\theta_{s}, eb_s))$}
\label{eq:final_loss}
\end{equation}

\section{Experiments}

For evaluation, we finetune the student model just as one would finetune the original BERT model i.e., without using the teacher model or any task-specific distillation. We describe our experiments below, with dataset details left to the appendix.

\subsection{Evaluation Tasks and Datasets}

We fine-tune and evaluate the distilled student models on two classes of language understanding tasks:
\\[7pt]
\noindent\textbf{GLUE benchmark} \citep{wang2018glue}\textbf{:} We pick three classification tasks from GLUE:
\begin{itemize}[leftmargin=*,topsep=2pt,itemsep=0pt]
\item MRPC: Microsoft Research Paraphrase Corpus \citep{dolan2005automatically}, a 2-way sentence pair classification task with 3.7K train instances.
\item MNLI: Multi-Genre Natural Language Inference \citep{williams2018broad}, a 3-way sentence pair classification task with 393K training instances.
\item SST-2: Stanford Sentiment Treebank \citep{socher2013recursive}, a 2-way sentence classification task with 67K training instances.
\end{itemize}
% \\[7pt]
\noindent\textbf{Spoken Language Understanding:} Since we are also keen on edge device applications, we also evaluate on spoken language understanding, a practical task in dialogue systems. We use the SNIPS dataset \citep{coucke2018snips} of $\sim$14K virtual assistant queries, each comprising one of 7 intents and values for one or more of the 39 pre-defined slots. The intent detection and slot filling subtasks are modeled respectively as 7-way sentence classification and sequence tagging with IOB slot labels. %\citep{sang2000introduction}
% Figure \ref{fig:clu_example} contains an example from the dataset.
% \begin{figure*}[h!]

% \setlength{\abovecaptionskip}{5pt}
% \setlength{\belowcaptionskip}{-5pt}
% \newcommand{\smalldownarrow}{\small$\downarrow$\normalsize}
% \centering
%   \begin{tabular}{l c c c c c}
%   \textit{intent} & SearchCreativeWork\\
% \textit{utterance} & Find & a & novel & called & industry\\
% & \smalldownarrow & \smalldownarrow & \smalldownarrow & \smalldownarrow & \smalldownarrow \\
%  \textit{slots}&  O & O & B-object\_type & O & B-object\_name\\
% \end{tabular}
%   \caption{Example from SNIPS dataset with
%   intent and $IOB$ slot annotations.}
%   \label{fig:clu_example}
% \end{figure*}

\subsection{Models and Baselines}

\begin{SCtable*} [0.4][t!]
% \centering
\setlength\tabcolsep{2.7pt}
\def\arraystretch{1.2}
\resizebox{0.73\textwidth}{!}
{

\begin{tabular}{l|c|ccc|c}
\toprule
% 'mnlimm' 'rte' 'qqp' 'qnli' '
% \rowcolor{lightgray}
\textbf{Model} & \textbf{\#Params} & \textbf{MRPC}   & \textbf{MNLI-m/mm}  & \textbf{SST-2} & \textbf{Average} \\ 
& & \textbf{(F1/Acc)} & \textbf{(Acc)} & \textbf{(Acc)} & \textbf{(F1/Acc)}\\
% $^{\ddag}$
\hline\hline
BERT\textsubscript{BASE} \cite{devlin2018bert} & 109M & $\,$ 88.9/-  & 84.6/83.4 &  93.5 & 89.0\\
BERT\textsubscript{LARGE} \cite{devlin2018bert} & 
340M & $\,$ 89.3/- & 86.7/85.9 &  94.9 & 90.3\\
\hline
PKD\textsubscript{6} \citep{sun2019patient} & 67.0M  & 85.0/79.9 & 81.5/81.0 & 92.0 & 86.2\\
PKD\textsubscript{3} \citep{sun2019patient} & 45.7M  & 80.7/72.5 & 76.7/76.3 & 87.5 & 81.6\\
% DistilBERT\textsubscript{6} \cite{sanh2019distilbert}\\
DistilBERT\textsubscript{4} \cite{sanh2019distilbert} & 52.2M & 82.4/- & 78.9/78.0 & 91.4 & 84.2\\
MobileBERT \citep{mobilebert} & 25.3M & 88.8/84.5 & 83.3/82.6 & 92.8 & 88.3\\
TinyBERT\textsubscript{4} \citep{jiao2019tinybert} & 14.5M & 82.0*/ - & 76.6/77.2* & - & -\\
TinyBERT\textsubscript{4}$^{\dag}$ \citep{jiao2019tinybert} & 14.5M & 86.4/- & 82.5/81.8  & 92.6 & 87.2\\
% BERT-of-Theseus\textsubscript{6}$^{\dag}$ \citep{xu2020bert} & 66M & - & 82.1*/- & 91.8* & -\\
BERT-of-Theseus\textsubscript{6}$^{\dag}$ \citep{xu2020bert} & 66M & 87.6/83.2 & 82.4/82.1 & 92.2 & 87.4\\

% BERT-of-Theseus\textsubscript{6} \citep{xu2020bert} & 66M & 87.6/83.2 & 82.4 / 82.1 & 92.2 & 87.4\\
\hline\hline

% NoKD Baseline, L-6, H-192 & \multirow{2}{*}{\shortstack{11.2M}}  & 0/0 & 0/0 & 0 & 0 \\ 
% DualTrain, L-6, H-192 & &   0/0 & 0/0 & 0 & 0 \\
\hline
NoKD Baseline, L-6, H-256 & 
\multirow{2}{*}{\shortstack{6.2M}}  & 81.2/74.1 & 76.9/76.1 & 87.0 & 81.7 \\ 
{\color{black}Mixed-vocab distilled} (ours), L-6, H-256 & & 84.9/79.3 & 79.0/78.6 & 89.1 & 84.3 \\
\hline
% NoKD Baseline, L-12, H-192 & \multirow{2}{*}{\shortstack{{19.3M}}}  & 82.6/74.3 & 77.4/76.5 & 87.1 & 0\\ 
% MixedDistill (ours), L-12, H-192 &  & 83.9/77.5 & 78.1/77.8 & 88.3 & 0 \\ \hline
NoKD Baseline, L-12, H-256 & 
\multirow{2}{*}{\shortstack{10.9M}}  & 85.1/79.8 & 79.1/79.0 & 89.4 & 84.5 \\ 
{\color{black}Mixed-vocab distilled} (ours), L-12, H-256 & &  87.2/82.6 & 80.7/80.5 & 90.6 & 86.2 \\
\toprule
\multicolumn{1}{l}{* denotes metrics on the development set} & \multicolumn{5}{r}{
$^{\dag}$ denotes results with task-specific distillation}\\
% \multicolumn{5}{l}{$^{\ddag}$ average is computed using F1 score for MRPC and accuracy for MNLI-m and SST-2}
\end{tabular}
}

\caption{Test set accuracy of distilled models, teacher model and baselines on the GLUE test sets, along with other parameters. MNLI-m and MNLI-mm refer to the genre-matched and mismatched test sets. All models other than \textit{NoKD} and our distilled models use a 30K-WordPiece vocabulary. The average is computed using F1 score for MRPC and accuracy for MNLI-m and SST-2.}
\label{tab:bert_pref_glue}
\end{SCtable*}

For GLUE, we train student models with 6 and 12 layers, 4 attention heads, and embedding/hidden dimensions fixed to 256, each using a compact 5K-WP vocabulary. We also evaluate baselines without knowledge distillation (\textit{NoKD}), parameterized identically to the distilled student models (incl. the 5K vocabulary), trained on the MLM teacher objective from scratch. We also compare our models on GLUE with the following approaches:

\setlist{nolistsep}

\begin{itemize}[leftmargin=*,topsep=2pt,itemsep=2pt]
    \item \textit{DistilBERT} \cite{sanh2019distilbert} distill BERT\textsubscript{BASE} to 4/6 layers by aligning teacher predictions,
    \item \textit{Patient KD - PKD} \cite{sun2019patient} align hidden states to distill BERT\textsubscript{BASE} to 3/6 layers,
    \item \textit{BERT-of-Theseus} \cite{xu2020bert} use a layer dropout method to distill BERT\textsubscript{BASE} to 6 layers,
    \item \textit{TinyBERT} \citep{jiao2019tinybert} apply task specific distillation to BERT\textsubscript{BASE} and align teacher outputs, hidden states as well as embeddings, and
    \item \textit{MobileBERT} \cite{mobilebert} combine layerwise transfer, architecture search and bottleneck structures for an optimized student model.
\end{itemize}

For SNIPS, we shift our focus to smaller, low-latency models for on-device use cases. Here, we train student models with 6 layers and embedding/hidden dimensions ${\in}\, \{96, 192, 256\}$. The smaller models here may not be competitive on GLUE but are adequate for practical tasks such as spoken LU. We compare with two strong baselines:

\begin{itemize}[leftmargin=*,topsep=2pt,itemsep=2pt]
\item BERT\textsubscript{BASE} \cite{chen2019bert} with intent and IOB slot tags predicted using the \texttt{[CLS]} and the first WP tokens of each word respectively, and
\item StackProp \cite{qin2019stack}, which uses a series of smaller recurrent and self-attentive encoders.
\end{itemize}

\subsection{Training Details}
\noindent\textbf{Distillation:} For all our models, we train the teacher model with mixed-vocabulary inputs (stage \RNum{1}) for 500K steps, followed by 300K steps of training just the student model (stage \RNum{2}). We utilize the same corpora as the teacher model i.e. BooksCorpus \citep{zhu2015aligning} and English Wikipedia.

For both stages, up to 20 input tokens were masked for MLM. In stage \RNum{1}, up to 10 of these masked tokens were tokenized using the teacher vocabulary, the rest using the student vocabulary.

We optimize the loss using LAMB \citep{you2019large} with a max learning rate of .00125, linear warmup for the first 10\% of steps, batch size of 2048 and sequence length of 128. Distillation was done on Cloud TPUs in a 8x8 pod configuration. $p_{SV}$, the probability of segmenting a Stage \RNum{1} input word using the student vocabulary, is set to 0.5.

%\footnote{https://cloudplatform.googleblog.com/2018/06/Cloud-TPU-now-offers-preemptible-pricing-and-global-availability.html}

\noindent\textbf{Finetuning:} For all downstream task evaluations on GLUE, we finetune for 10 epochs using LAMB with a learning rate of 0.0001 and batch size of 64. For all experiments on SNIPS, we use ADAM with a learning rate of 0.0001 and a batch size of 64.

\definecolor{mygray}{gray}{0.93}

\section{Results}

\noindent\textbf{GLUE:} Table \ref{tab:bert_pref_glue} shows results on the downstream GLUE tasks alongside model sizes, for our models, BERT\textsubscript{BASE/LARGE}, and all baselines.

Our proposed models consistently improve upon the identically parameterized \textit{NoKD} baselines, indicating that mixed-vocabulary training is better than training from scratch and avoids a drastic drop in performance from the teacher. Compared with PKD and DistilBERT, our 6-layer model outperforms PKD\textsubscript{3} while being $>$7x smaller and our 12-layer model performs competitively with PKD\textsubscript{6} and DistilBERT\textsubscript{4} while being $\sim$5-6x smaller.

Interestingly, our models do particularly well on the MRPC task: the 6-layer distilled model performs almost as well as PKD\textsubscript{6} while being over 10x smaller. This may be due to our smaller models being data-efficient on the smaller MRPC dataset.

{\color{black}TinyBERT and Bert-of-Theseus are trained in task-specific fashion i.e., a teacher model already finetuned on the downstream task is used for distillation. TinyBERT's non-task-specific model results are reported on GLUE dev sets: these results are, therefore, not directly comparable with ours. Even so, our 12-layer model performs credibly compared with the two, presenting a competitive size-accuracy tradeoff, particularly when compared to the 6x larger BERT-of-Theseus.

MobileBERT performs strongly for the size while being task-agnostic. Our 12-layer model, in comparison, retains $\sim$98\% of its performance with 57\% fewer parameters and may thus be better-suited for use on highly resource-limited devices.

TinyBERT sees major gains from task-specific data augmentation and distillation, and MobileBERT from student architecture search and bottleneck layers. Notably, our technique targets the student vocabulary without conflicting with any of the above methods and can, in fact, be combined with these methods for even smaller models.}\\[7pt]\noindent\textbf{SNIPS:} Table \ref{tab:snips} shows results on the SNIPS intent and slot tasks for our models and two state-of-the-art baselines. Our smallest 6-layer model retains over 95\% of the BERT\textsubscript{BASE} model's slot filling F1 score \citep{sang2000introduction} while being 30x smaller (${<}$ 10 MB w/o quantization) and 57x faster on a mobile device, yet task-agnostic. Our other larger distilled models also demonstrate strong performance ($0.2{\text -}0.5\%$ slot F1 higher than the respective \textit{NoKD} baselines) with small model sizes and latencies low enough for real-time inference. This indicates that small multi-task BERT models \cite{tsai2019small} present better trade-offs for on-device usage for size, accuracy and latency versus recurrent encoder-based models such as StackProp.

% While we experience a larger performance drop for GLUE benchmark tasks, it is worth noting that these are purpose-built to capture general natural language understanding capabilities. Real-world use cases such as spoken language understanding, in contrast, may still be served successfully using smaller language models as presented.

\begin{table}[t!]
\centering
\setlength\tabcolsep{3.3pt}
\def\arraystretch{1.25}
\resizebox{\columnwidth}{!}{

\begin{tabular}{l|cc|cc}
\toprule
% 'mnlimm' 'rte' 'qqp' 'qnli' '
% \rowcolor{lightgray}
\textbf{Model} & \textbf{\#Params} & \textbf{Latency} & \textbf{Intent Acc} & \textbf{Slot F1} \\
\hline\hline
BERT\textsubscript{BASE} \cite{chen2019bert} & 109M & 340ms & 98.6 & 97.0\\
StackProp \cite{qin2019stack} & 2.6M & ${>}$70ms & 98.0 & 94.2\\\hline
{\color{black}Mixed-vocab distilled}, L-6, H-96 & 1.2M & 6ms & 98.9 & 92.8\\
% DualTrain, L-12, H-96 & 5.6M & 14ms & 98.9 & 92.8\\
{\color{black}Mixed-vocab distilled}, L-6, H-192 & 3.6M & 14ms & 98.8 & 94.6\\
{\color{black}Mixed-vocab distilled}, L-6, H-256 & 6.2M & 20ms & 98.7 & 95.0\\
\toprule
\end{tabular}
}

\caption{Results on the SNIPS dataset. Latency is measured with 4 CPU threads on a {\color{black}Pixel 4 mobile device}.}
\label{tab:snips}
\end{table}

\section{Discussion}
\label{section:discussion}
\noindent\textbf{Impact of vocabulary size:} We trained a model from scratch identical to BERT\textsubscript{BASE} except with our 5K-WP student vocabulary. On the SST-2 and MNLI-m dev sets, this model obtained 90.9\% and 83.7\% accuracy respectively - only 1.8\% and 0.7\% lower respectively compared to BERT\textsubscript{BASE}.

Since embeddings account for a larger fraction of model parameters with fewer layers, we trained another model identical to our 6$\times$256 model, but with a 30K-WP vocabulary and teacher label distillation. This model showed small gains (0.1\% / 0.5\% accuracy on SST-2 / MNLI-m dev) over our analogous distilled model, but with $30\%$ more parameters solely due to the larger vocabulary. 

This suggests that a small WordPiece vocabulary may be almost as effective for sequence classification/tagging tasks, especially for smaller BERT models and up to moderately long inputs. Curiously, increasing the student vocabulary size to 7K or 10K did not lead to an increase in performance on GLUE. We surmise that this may be due to underfitting owing to the embeddings accounting for a larger proportion of the model parameters.
\\[7pt]
{\color{black}\noindent\textbf{Alternative vocabulary pruning:} Probing other strategies for a small-vocabulary model, we used the above 6$\times$256 30K-WP vanilla distilled model to obtain a smaller model by pruning the vocabulary to contain the intersection of the 30K and 5K vocabularies (total 4629 WPs). This model is 1.2\% smaller than our 4928-WP distilled model, but drops 0.8\% / 0.7\% on SST-2/MNLI-m dev sets.

Furthermore, to exploit the high overlap in vocabularies, we tried running our distillation pipeline but with the embeddings for student tokens (after projecting up to the teacher dimension) also present in the teacher vocabulary tied to the teacher embeddings for those tokens. This model, however, dropped 0.7\% / 0.5\% on SST-2/MNLI-m compared to our analogous 6$\times$256 distilled model.

We also tried pretraining BERT\textsubscript{LARGE} from scratch with the 5K vocabulary and doing vanilla distillation for a 6$\times$256 student: this model dropped 1.2\% / 0.7\% for SST-2/MNLI-m over our similar distilled model, indicating the efficacy of mixed-vocabulary training over vanilla distillation.}
%, likely since teacher embeddings were not used for transfer.
\\[7pt]

\section{Conclusion}

We propose a novel approach to knowledge distillation for BERT, focusing on using a significantly smaller vocabulary for the student BERT models. Our \textit{mixed-vocabulary training} method encourages implicit alignment of the teacher and student WordPiece embeddings. Our highly-compressed 6 and 12-layer distilled student models are optimized for on-device use cases and demonstrate competitive performance on both benchmark datasets and practical tasks. Our technique is unique in targeting the student vocabulary size, enabling easy combination with most BERT distillation methods.

\bibliography{eacl2021}
\bibliographystyle{acl_natbib}
\end{document}